\documentclass[11pt,twocolumn]{article}
\usepackage[utf8]{inputenc}

  \usepackage[
    backend=biber,
    style=numeric,
  ]{biblatex}

\usepackage{pgfplots} 

\title{Defining maximum acceptable latency of AI-enhanced CAI tools}
\author{Claudio Fantinuoli\textsuperscript{1,2}, Maddalena Montecchio\textsuperscript{1}\\
  \textsuperscript{1} Johannes Gutenberg University Mainz\\
  \textsuperscript{2} KUDO
  }
\date{}

\begin{document}

\maketitle

\section{Abstract}
Recent years have seen an increasing number of studies around the design of computer-assisted interpreting tools with integrated automatic speech processing and their use by trainees and professional interpreters. This paper discusses the role of system latency of such tools and presents the results of an experiment designed to investigate the maximum system latency that is cognitively acceptable for interpreters working in the simultaneous modality. The results show that interpreters can cope with a system latency of 3 seconds without any major impact in the rendition of the original text, both in terms of accuracy and fluency. This value is above the typical latency of available AI-based CAI tools and paves the way to experiment with larger context-based language models and higher latencies.

\section{Introduction}
Automatic speech recognition (ASR) has been regarded as a technology ``with considerable potential for changing the way interpreting is practiced'' \cite{pochhacker_introducing_2016}. In particular, ASR has been proposed as a means to overcome the shortcomings in current implementations of computer-assisted interpreting (CAI) tools, such as the inherent difficulty to manually lookup terms while interpreting. By integrating AI-based features in the interpreter workstation, such as the real-time automatic suggestion of numbers and other problem triggers \cite{fantinuoli_speech_2017}, and by integrating advancements in extractive and predictive algorithms based on machine learning \cite{vogler_lost_2019}, the cognitive load required by the use of CAI tools may be reduced, offering new opportunities for interpreters to improve their performance. First empirical research seems to point in this direction \cite{defrancq_automatic_2020, fantinuoli_measuring_2021}.

AI-enhanced CAI tools are generally based on the concatenation of several modules that may comprise a speech-to-text transcription engine, a parsing module to identify the units of interest, for example numbers, terminology or proper names, and a visualisation component for the human-machine interaction. Other implementations may adopt an end-to-end approach, making the cascading architecture obsolete. Such tools can be deployed on the edge, running completely on the user's device, or on the web, hosted on servers and accessed through a Web Browser. Generally speaking, CAI tools present suggestions with a certain amount of delay with respect to the original speech. This delay is due to the architectural design of the tool and its components, for example the ASR latency or the computation time needed to inference a language model, but also to the intrinsic nature of the task, which may require a certain amount of linguistic context in order to make an informed decision on what to show to the final user. In a time-sensitive activity such as simultaneous interpreting, system latency may require interpreters to adopt new interpreting strategies to successfully integrate the suggestions into their rendition. If latency, however, exceeds a certain threshold, interpreters may not be able to integrate the suggestions, or may do this at the expenses of fluency, cohesion, or even accuracy. In this case, the use of such high-latency tool would become detrimental for the user-machine interaction, for the interpreter's cognitive load, and finally for the quality of the rendition. 

Not much is known about this maximum acceptable latency threshold, i.e. the maximum ear-voice-span (EVS) that interpreters can cope with in order to successfully integrate external suggestions without impacting the overall interpreting performance. Not only this value may vary among subjects, but it may also be different from the one reported in experimental analysis of interpretation performed without the use of a CAI tool. 

Our hypothesis is that displaying suggestions within the average EVS reported in literature should not negatively impact the interpreter rendition, even when this is `forced' upon the interpreter by the latency with which the tool generates such suggestions. To our knowledge this is the first study that tries to empirically answer this question. On the one hand, knowledge about this threshold is crucial to understand if current AI-based CAI implementations can already meet interpreters expectations or if major efforts should be placed in further reducing system latency. On the other hand, an acceptable higher latency could allow the integration of more time-demanding NLP features, such as the automatic prediction of difficult parts of the speech \cite[eg.][]{vogler_lost_2019}.

The reminder of this paper is organized as follows. Section 3 introduces the related work in the area of CAI tools and the empirical experiments conducted on them so far. Section 4 describes the data and the methodology adopted in this experiment. Section 5 introduces the evaluation framework. Section 6 presents the results. Finally, section 7 discusses the conclusion and the outlook.

\section{Related work}

Recent years have seen an increasing number of studies around the design of AI-based computer-assisted interpreting tools and their use by trainees and professional interpreters. 

Computer-assisted interpreting tools are digital devices designed to support the interpreter in different steps of their work, from preparation to the very act of interpreting. They have been proposed by several researchers in the past 20 years or so \cite{will_zur_2015,rutten_terminology_2017,stoll_jenseits_2009}, but it is only recently that there has been a surge in interest among the community's members. 

Thanks to new advances in artificial intelligence, ASR has reached a quality level that makes it suitable for integration into supportive technologies. By automating and extending the query system of CAI tools, this integration may solve the shortcomings of traditional tools \cite[eg.][]{fantinuoli_interpretbank._2016, fantinuoli_computer-assisted_2017, hansen-schirra_nutzbarkeit_2012} and extend the features available in an interpreters’ workstation, for example  automatically suggesting translations of specific terms as well as transcribing numbers and proper names in real-time. 

Over the years, a handful of empirical studies have been carried out to test the feasibility of the human-machine interaction in the simultaneous modality. They have focused in particular on the effectiveness of ASR-support during the interpretation of numbers \cite{desmet_simultaneous_2018, defrancq_automatic_2020, fantinuoli_measuring_2021}, one of the problem triggers of simultaneous interpreting identified in literature \cite{braun_inaccuracy_1996, frittella_706_2019, gile_basic_2009, setton_conference_2016}. In order to measure the impact on the quality of the rendition, these studies have used either mock-up systems with a very short latency \cite{desmet_simultaneous_2018, canali_technologie_2019} or real-life tools with a reported latency of under 2 seconds \cite{defrancq_automatic_2020, fantinuoli_measuring_2021}.  

From an interpreter perspective, the system latency impacts the ear-voice span (EVS) of interpreters since it forces the interpreter to wait a certain amount of time before being able to integrate the suggestions into their rendition.  The EVS is the amount of time that separates the words uttered by the speaker in the source speech and the equivalent rendition in the target speech uttered by the interpreter \cite{pochhacker_introducing_2016}. Research in this field has a “long-standing tradition” \cite{defrancq_corpus-based_2015} and EVS is considered an essential variable that can potentially impact interpreters’ performance \cite{barik_simultaneous_1975, gerver_effects_1969, gile_basic_2009, lee_ear_2004} such that it continues to be an object for analysis and assessment. 

Previous research has found that the variable limits of interpreters’ EVS are generally attributable to two main factors: input-related factors, which might also enable interpreters to consciously regulate their lag, and personal factors, e.g. short term memory-related factors, that influence the maximum capacity of the memory \cite{setton_conference_2016, anderson_simultaneous_1994, lederer_simultaneous_1978}. Interpreters are free to regulate - at least to a certain degree - their EVS depending on the features of the speech segment they are translating, increasing or decreasing it consciously as a strategy \cite{anderson_simultaneous_1994, donato_strategies_2003}. 

Measuring EVS is not immediate and easy, due to the fact that grammar, word order and syntactic structure might be different in the two languages \cite{setton_conference_2016}. Average EVS reported in literature is between 2 and 3 seconds \cite{barik_simultaneous_1975, lederer_simultaneous_1978, lee_ear_2004}, with a peak of 10 seconds \cite{oleron_research_2001}; if measured in words, a mean ranging from 5 to 10 words has been reported \cite{schweda_linguistic_1987} .

\section{Data and methodology}

\subsection{Dataset}\label{dataset}

For this experiment we used numbers as unit of interest to be suggested by the system, as this is a problem trigger that has been largely studied so far in empirical experiments related to CAI tools. We choose and edited therefore a speech particularly dense with numbers. The speech had a duration of 7 minutes and 15 seconds and did not pose any particular difficulties in terms of terminology.  

The speech was delivered in English by a non-native speaker and participants were asked to translate it simultaneously into their native language (Italian). The average speech pace was 105.5 words per minute, which corresponds to an ideal speech rate for the simultaneous interpretation of improvised speeches and which is close to the ideal read-aloud speech rate of 100 words per minute. The speech had 25 stimuli (numbers); 10 of them were accompanied by a referent. 

The speech was prerecorded and a video simulating an ASR system was edited ad-hoc by the authors of the experiment in order to retain full control over the variable `latency'. In the video the transcription of the numbers (without the relevant referent) was shown on screen after a variable delay compared to the moment the participants received the acoustic signal in the headset. The acoustic numerical input was removed from the recorded speech and replaced with a neutral acoustic signal (/beep/). This choice was made in order to force the participants to use the visualization of the numbers on screen, thus relying exclusively on the ASR simulation. Every visual input (number) was displayed based on a preset latency, which gradually increased during the course of the speech. Numbers were shown in isolation with no embedded transcription of the speech. The text was divided into five different sections. In every section the visualization of the stimulus occurred with a different latency – in the first section, the numbers appeared on the screen 1 second after the acoustic signal, in the second section after 2 seconds, and so forth. During their performance, participants received the visual numerical input on a maxi screen situated inside the classroom or on the monitors inside the booths. 

\subsection{Participants}

A total of eight participants took part in this study. All participants were students. They were Italian native speakers with German and English in their language combination enrolled in their final year of a Master in Conference Interpreting program and had at least 1 year experience in simultaneous interpretation. The participants did not have past experience of interpreting with CAI tools or ASR-enhanced CAI tools. This condition is in line with experimental setups adopted in similar studies \cite[eg.][]{defrancq_automatic_2020, fantinuoli_measuring_2021}.  

Some weeks before the experiment, the participants were provided with three short videos containing a simulation of the ASR system in order to be minimally exposed to the experimental format. These videos were edited along the lines of the video used for the actual experiment, but all numerical visual inputs were displayed with the exact same latency, in contrast with the variable latency with which the numerical visual inputs were displayed in the experiment. The goal was to help the participants to get accustomed to the experiment format, i.e., the way numbers were transcribed, their font size, the color and the way the order of magnitude was shown \cite{fantinuoli_measuring_2021}. 

\section{Evaluation framework and procedure}

The analysis of the collected data has been performed on two different levels, namely on a stimuli-based and on a segment-based level. 

The aim of the \textit{stimuli-based evaluation} is to assess the accuracy level achieved by the interpreters in rendering the units of interest in the target language, depending on the variable latency of the mockup system. Within this level of assessment, the accurate rendition of the units of interest is evaluated considering the following components: the numerical information comprising its order of magnitude, and, if present, the relevant referent. A number of parameters were collected for the evaluation: presence of the number in the rendition, accuracy of the number, presence of the referent, accuracy of the referent, pronunciation disfluency. Numbers and referents that were approximated, generalized or omitted were considered to be errors. Moreover, lexical, syntactic, phonological or articulation mistakes were also classified as errors. 

The \textit{segment-based evaluation} assesses the quality of the entire segments in which the numbers are embedded. This evaluation focuses on two different parameters. The first is the accuracy of the target segment, measured in terms of the following linguistic aspects: faithfulness, grammatical correctness, completeness, logical cohesion, consistency, plausibility. The second parameter is the listener’s perception in terms of delivery flow. This evaluation level focuses on the following paralinguistic aspects: perception of the interpreters’ voice and rhythm, eloquence, presentation, prosody and communicative effectiveness. The segment-based evaluation has been performed using a Likert scale (1 to 5) by three different evaluators. 

There are three major limitations in this study that could be addressed in future research. Firstly, the level of interpretation proficiency of the participants selected for this experiment is biased towards the lower end. It is reasonable to assume that a randomized sample in terms of proficiency, for example including also professional interpreters, may increase the maximum acceptable latency measured in the experiment. Secondly, the experiment focused only on a single language pair. Because of intrinsic variations among languages, EVS and different strategies adopted by interpreters on the basis of the language combination may influence the threshold under scrutiny. Finally, variations in speech complexity, for example in terms of syntactic structures, delivery speed etc.  have not been taken into consideration in our study. However, they may have a considerable impact on interpreters cognitive load and consequently on their ability to successfully integrate suggestions and may require a dynamic adaptation of latency according to this variable.

\section{Results}

\subsection{Stimuli-based evaluation}

As introduced in the previous section, the stimuli-based evaluation focuses on the stimuli (numbers) and their referents and not on the whole sentence in which they are embedded. 

\begin{figure}
\centering
\begin{tikzpicture}
\begin{axis} [ybar,bar width=15pt]
\addplot coordinates {
    (1,97.14) 
    (2,97.14) 
    (3,98.86) 
    (4,93.14)
    (5,94.86)
};
\end{axis}
\end{tikzpicture}
\caption{Accuracy of number rendition} \label{fig:accuracy_rendition}
\end{figure}
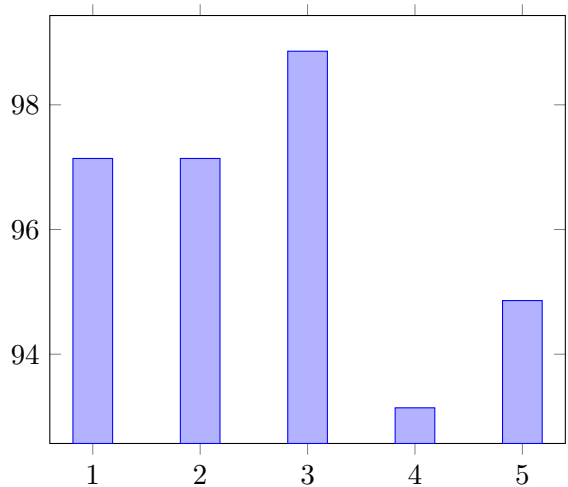

No number and no referent has been omitted at any latency by any of the candidates. The precision, however, varies according to the latency. Figure \ref{fig:accuracy_rendition} presents the accuracy of the rendition of the numbers displayed by the ASR simulation system. The best results were achieved with the first three latencies: 97.14\% at 1 and 2 seconds and 98.85\% at 3 seconds. The better score at 3 seconds may be explained with the interpreters acclimatisation to the dynamic of the experiment, thus enabling them to adapt their interpreting approach after having generated a learning effect \cite{mellinger_interpreter_2018}. The worst results were observed when the latency increased above 3 seconds, and in particular 93.14\% at 4 and 94.86\% at 5 seconds. 

In Figure \ref{fig:reference_rendition} the accuracy of the referents of the displayed numbers is presented. The highest score was reached with a latency of 3 seconds (100\%). The accuracy decreases with the two higher latencies: 94.28\% at 4 seconds and 85.71\% at five seconds. 

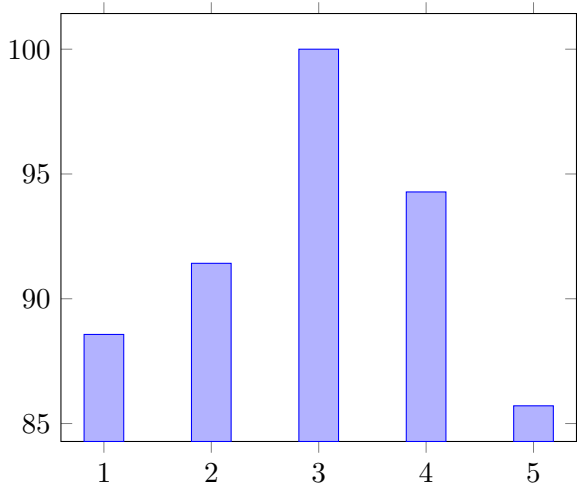
\begin{figure}
\centering
\begin{tikzpicture}
\begin{axis} [ybar,bar width=15pt]
\addplot coordinates {
    (1,88.57) 
    (2,91.42) 
    (3,100) 
    (4,94.28)
    (5,85.71)
};
\end{axis}
\end{tikzpicture}
\caption{Accuracy of reference rendition} \label{fig:reference_rendition}
\end{figure}

\begin{figure}
\centering
\begin{tikzpicture}
\begin{axis} [ybar,bar width=15pt]
\addplot coordinates {
    (1,2.85) 
    (2,5.71) 
    (3,5.71) 
    (4,25.71)
    (5,8.57)
};
\end{axis}
\end{tikzpicture}
\caption{Disfluency in number rendition} \label{fig:disfluency}
\end{figure}
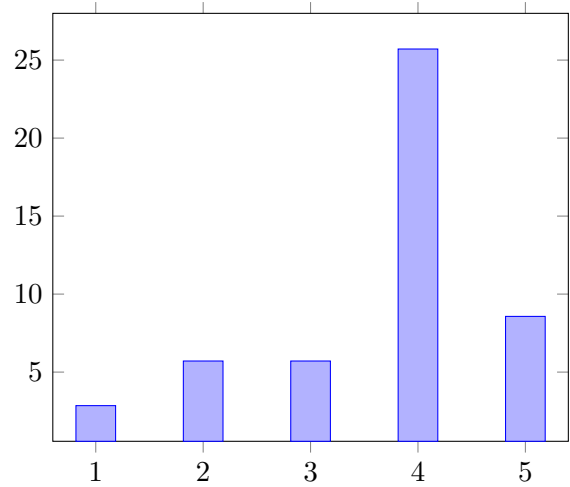

The number of pronunciation disfluencies (Figure \ref{fig:disfluency}) reached a pick with a latency of 4 seconds (25.71\% of pronounced numbers present a disfluency). With lower latencies the number of disfluencies is low: 2.85\% of numbers were disfluent at 1 second and 5.71\% at 2 and 3 seconds.

\subsection{Segment-based evaluation}

The segment-based evaluation aims at extending the assessment of the quality of the rendition from the unit of interest to the entire segment in which this unit appears. The segment-level accuracy was higher with a latency of 1 and 2 seconds. In particular, the best score was reached with 2 seconds. As in the stimuli-based evaluation, this leads us to conclude that the participants, although able to successfully integrate the accurate interpretation of the numbers displayed in their own EVS, needed a settling period in order to gain familiarity with the speech and the experimental format. 

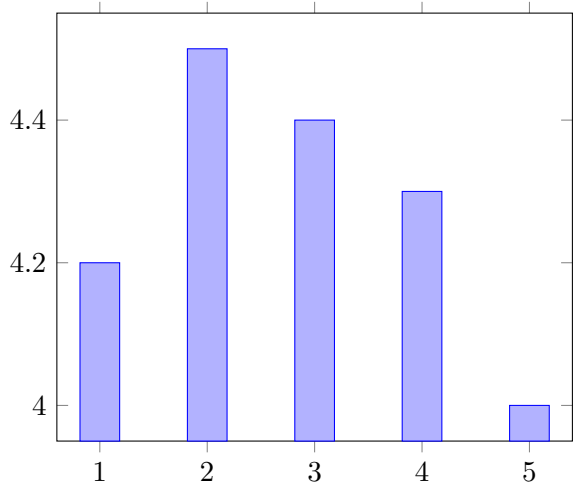
\begin{figure}
\centering
\begin{tikzpicture}
\begin{axis} [ybar,bar width=15pt]
\addplot coordinates {
    (1,4.2) 
    (2,4.5) 
    (3,4.4) 
    (4,4.3)
    (5,4.0)
};
\end{axis}
\end{tikzpicture}
\caption{Accuracy at segment level} \label{fig:reference_rendition}
\end{figure}

A minimal decline in segment-level accuracy was observed from latency 3 onward,  while with 5 seconds latency the accuracy degrades quite considerably. It can be assumed that with this high latency, participants experienced difficulties in allocating the correct amount of cognitive resources, since they were required to focus on several tasks – on the screen with the number displayed with a specific delay, on retaining the information they had just heard in working memory and on processing the new segments of the speech that continued to be uttered by the speaker \cite[eg.][]{gile_basic_2009}. 

The fluency of the delivery was evaluated high in the latency range from 1 to 3 seconds. Paralinguistic aspects, e.g., the voice and the fluency of the rendition have been rated more agreeable in the first three sections. With 4 seconds, the delivery flow deteriorated significantly, and it reached its lowest score in the 5 seconds latency section. It is plausible to assume that not only the increasing cognitive load triggered by a longer EVS also had significant repercussions on the manner in which the interpreters delivered the target speech, as it adversely affected paralinguistic aspects, such as the rhythm, the intonation, the voice and the appearance \cite[eg.][]{buhler_linguistic_1986, becerra_first_2016}.  

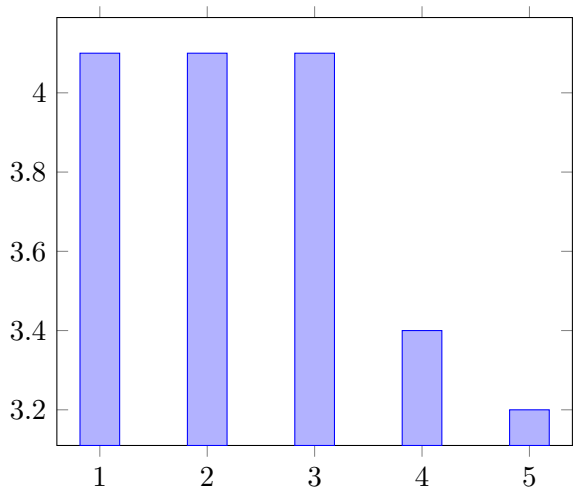
\begin{figure}
\centering
\begin{tikzpicture}
\begin{axis} [ybar,bar width=15pt]
\addplot coordinates {
    (1,4.1) 
    (2,4.1) 
    (3,4.1) 
    (4,3.4)
    (5,3.2)
};
\end{axis}
\end{tikzpicture}
\caption{Fluency at segment level} \label{fig:reference_rendition}
\end{figure}

\section{Conclusions and outlook}
In this paper we presented the results of an empirical experiment aimed at measuring the maximum acceptable latency of an automatic suggestion feature for simultaneous interpretation. The results seem to suggest that interpreters are able to integrate suggestions by ad-hoc extending their ear-voice-span to 2 seconds without compromising the quality of their rendition and to 3 seconds without any major disruption. A further extension of the system latency seems to induce a consistent reduction of precision in the use of such suggestions and in the emergence of information losses in the overall rendition. This is in line with our original hypothesis that the system latency should not exceed the average interpreter's EVS. Within the latency threshold outlined in the experiment, next generation AI-enhanced CAI tools could be able to accommodate more complex and context-based NLP-features without a significant risk of impairing the usability of the tool.

\printbibliography 

@book{gile_basic_2009,
	address = {Amsterdam},
	edition = {2nd},
	title = {Basic {Concepts} and {Models} for {Interpreter} and {Translator} {Training}: {Revised} edition},
	isbn = {978-90-272-2433-0 978-90-272-2432-3 978-90-272-8808-0},
	shorttitle = {Basic {Concepts} and {Models} for {Interpreter} and {Translator} {Training}},
	language = {en},
	urldate = {2015-09-16},
	publisher = {John Benjamins Publishing Company},
	author = {Gile, Daniel},
	month = nov,
	year = {2009},
}

@book{stoll_jenseits_2009,
	title = {Jenseits simultanfähiger {Terminologiesysteme}},
	publisher = {Trier: Wvt Wissenschaftlicher Verlag},
	author = {Stoll, Christoph},
	year = {2009},
	note = {bibtex: stoll\_jenseits\_2009},
}

@article{will_zur_2015,
	title = {Zur {Eignung} simultanfähiger {Terminologiesysteme} für das {Konferenzdolmetschen}},
	volume = {8},
	number = {1},
	journal = {trans-kom},
	author = {Will, Martin},
	year = {2015},
	pages = {179--201},
}

@inproceedings{fantinuoli_interpretbank._2016,
	address = {London},
	title = {{InterpretBank}. {Redefining} computer-assisted interpreting tools},
	booktitle = {Proceedings of the {Translating} and the {Computer} 38 {Conference}},
	publisher = {Editions Tradulex},
	author = {Fantinuoli, Claudio},
	year = {2016},
	pages = {42--52},
}

@article{fantinuoli_computer-assisted_2017,
	title = {Computer-assisted preparation in conference interpreting},
	volume = {9},
	number = {2},
	journal = {Translation \& Interpreting},
	author = {Fantinuoli, Claudio},
	year = {2017},
	pages = {24--37},
}

@article{hansen-schirra_nutzbarkeit_2012,
	title = {Nutzbarkeit von {Sprachtechnologien} für die {Translation}},
	volume = {5},
	number = {2},
	journal = {trans-kom},
	author = {Hansen-Schirra, Silvia},
	year = {2012},
	pages = {211--226},
}

@book{setton_conference_2016,
	address = {Amsterdam \& Philadelphia},
	series = {Benjamins translation library ({BTL})},
	title = {Conference interpreting: a complete course},
	isbn = {978-90-272-5861-8 978-90-272-5862-5},
	shorttitle = {Conference interpreting},
	number = {volume 120},
	publisher = {John Benjamins Publishing Company},
	author = {Setton, Robin and Dawrant, Andrew},
	year = {2016},
}

@book{pochhacker_introducing_2016,
	edition = {2nd},
	title = {Introducing {Interpreting} {Studies}},
	publisher = {Routledge},
	author = {Pöchhacker, Franz},
	year = {2016},
}

@inproceedings{fantinuoli_speech_2017,
	address = {London},
	title = {Speech {Recognition} in the {Interpreter} {Workstation}},
	booktitle = {Proceedings of the {Translating} and the {Computer} 39},
	publisher = {London},
	author = {Fantinuoli, Claudio},
	year = {2017},
}

@inproceedings{rutten_terminology_2017,
	address = {Geneva},
	title = {Terminology {Management} {Tools} for {Conference} {Interpreters} –  {Current} {Tools} and {How} {They} {Address} the {Specific} {Needs} of  {Interpreters}},
	booktitle = {Proceedings of the 39th {Conference} {Translating} and the {Computer}},
	publisher = {Tradulex},
	author = {Rütten, Anja},
	year = {2017},
	pages = {98--103},
}

@article{vogler_lost_2019,
	title = {Lost in {Interpretation}: {Predicting} {Untranslated} {Terminology} in {Simultaneous} {Interpretation}},
	shorttitle = {Lost in {Interpretation}},
	url = {http://arxiv.org/abs/1904.00930},
	language = {en},
	urldate = {2019-05-04},
	journal = {arXiv:1904.00930 [cs]},
	author = {Vogler, Nikolai and Stewart, Craig and Neubig, Graham},
	month = apr,
	year = {2019},
	note = {arXiv: 1904.00930},
}

@article{mellinger_interpreter_2018,
	title = {Interpreter traits and the relationship with technology and visibility},
	volume = {13},
	issn = {1932-2798, 1876-2700},
	doi = {10.1075/tis.00021.mel},
	language = {en},
	number = {3},
	urldate = {2020-01-14},
	journal = {Translation and Interpreting Studies},
	author = {Mellinger, Christopher D. and Hanson, Thomas A.},
	month = nov,
	year = {2018},
	pages = {366--392},
}

@incollection{desmet_simultaneous_2018,
	title = {Simultaneous interpretation of numbers and the impact of technological support},
	booktitle = {Interpreting and technology, {Language} {Science} {Press}},
	publisher = {Language Science PRess},
	author = {Desmet, Bart and Vandierendonck, Mieke and Defrancq, Bart},
	year = {2018},
	pages = {13--27},
}

@phdthesis{canali_technologie_2019,
	address = {Roma},
	title = {Technologie und {Zahlen} beim {Simultandolmetschen}: utilizzo del riconoscimento vocale come supporto durante l’interpretazione simultanea dei numeri},
	school = {Università degli studi internazionali di Roma},
	author = {Canali, Sara},
	year = {2019},
}

@article{frittella_706_2019,
	title = {“70.6 {Billion} {World} {Citizens}”: {Investigating} the difficulty of interpreting number},
	volume = {11},
	number = {1},
	journal = {Translation \& Interpreting},
	author = {Frittella, Francesca},
	year = {2019},
	pages = {79--99},
}

@article{defrancq_automatic_2020,
	title = {Automatic speech recognition in the booth: {Assessment} of system performance, interpreters’ performances and interactions in the context of numbers},
	issn = {0924-1884, 1569-9986},
	shorttitle = {Automatic speech recognition in the booth},
	doi = {10.1075/target.19166.def},
	language = {en},
	urldate = {2020-11-29},
	journal = {Target. International Journal of Translation Studies},
	author = {Defrancq, Bart and Fantinuoli, Claudio},
	month = nov,
	year = {2020},
}

@incollection{fantinuoli_measuring_2021,
	title = {Measuring the impact of automatic speech recognition on interpreter’s performances in simultaneous interpreting},
	booktitle = {Empirical studies of translation and interpreting: the post-structuralist apporach},
	publisher = {Routledge},
	author = {Fantinuoli, Claudio and Pisani, Elisabetta},
	editor = {Caiwen, Wang and Binghan, Zheng},
	year = {2021},
}

@article{braun_inaccuracy_1996,
	title = {Inaccuracy for {Numerals} in {Simultaneous} {Interpretation}: {Neurolinguistic} and {Neuropsychological} {Perspectives}},
	volume = {7},
	journal = {The Interpreters’ Newsletter},
	author = {Braun, Sabine and Clarici, Andrea},
	year = {1996},
	pages = {85--102},
}

@article{defrancq_corpus-based_2015,
	title = {Corpus-based research into the presumed effects of short {EVS}},
	volume = {17},
	number = {1},
	journal = {Interpreting. International Journal of Research and Practice in Interpreting},
	author = {Defrancq, Bart},
	year = {2015},
	pages = {26--45},
}

@article{barik_simultaneous_1975,
	title = {Simultaneous {Interpretation}: {Qualitative} and {Linguistic} {Data}},
	volume = {18},
	number = {3},
	journal = {Language and Speech},
	author = {Barik, Henri},
	year = {1975},
	pages = {272--297},
}

@article{gerver_effects_1969,
	title = {Effects of {Grammaticalness}, {Presentation} {Rate}, and {Message} {Length} on {Auditory} {Short}-{Term} {Memory}},
	volume = {21},
	number = {3},
	journal = {Quarterly Journal of Experimental Psychology},
	author = {Gerver, David},
	year = {1969},
	pages = {203--208},
}

@article{lee_ear_2004,
	title = {Ear {Voice} {Span} in {English} into {Korean} {Simultaneous} {Interpretation}},
	volume = {47},
	number = {4},
	journal = {Meta},
	author = {Lee, Taehyung},
	year = {2004},
	pages = {596--606},
}

@incollection{anderson_simultaneous_1994,
	title = {Simultaneous {Interpretation}: {Contextual} and translation aspects},
	booktitle = {Bridging the {Gap}: {Empirical} {Research} in {Simultaneous} {Interpretation}},
	publisher = {Benjamins Translation Library},
	author = {Anderson, Linda and Lambert, Sylvie and Moser-Mercer, Barbara},
	year = {1994},
	pages = {101--120},
}

@article{donato_strategies_2003,
	title = {Strategies adopted by student interpreters in {SI}: a comparison between the {English}-{Italian} and the {German}-{Italian} language-pairs},
	volume = {12},
	journal = {Interpreter’s Newsletter},
	author = {Donato, Valentina},
	year = {2003},
}

@incollection{lederer_simultaneous_1978,
	title = {Simultaneous {Interpretation}—{Units} of {Meaning} and other {Features}},
	booktitle = {Language {Interpretation} and {Communication}},
	publisher = {Springer},
	author = {Lederer, Marianne},
	editor = {Gerver, David and Sinaiko, H. Wallace},
	year = {1978},
	pages = {323--332},
}

@incollection{oleron_research_2001,
	title = {Research into simultaneous translation},
	booktitle = {The interpreting studies reader},
	author = {Oléron, P. and Nanpon, H.},
	editor = {Pöchhacker, Franz and Shlesinger, Miriam},
	year = {2001},
	pages = {69--76},
}

@article{schweda_linguistic_1987,
	title = {Linguistic and {Extralinguistic} {Aspects} of {Simultaneous} {Interpretation}},
	volume = {8},
	journal = {Applied Linguistics},
	author = {Schweda, Nicholson},
	year = {1987},
	pages = {194--205},
}

@article{becerra_first_2016,
	title = {Do first impressions matter? {The} effect of first impressions on the assessment of the quality of simultaneous interpreting},
	volume = {17},
	number = {1},
	journal = {Across Languages and Cultures},
	author = {Becerra, Garcia},
	year = {2016},
	pages = {77--98},
}

@article{buhler_linguistic_1986,
	title = {Linguistic (semantic) and extra-linguistic (pragmatic) criteria for the evaluation of conference interpretation and interpreters},
	journal = {Multilingua},
	author = {Bühler, Hildegrund},
	year = {1986},
	pages = {231--235},
}

\end{document}